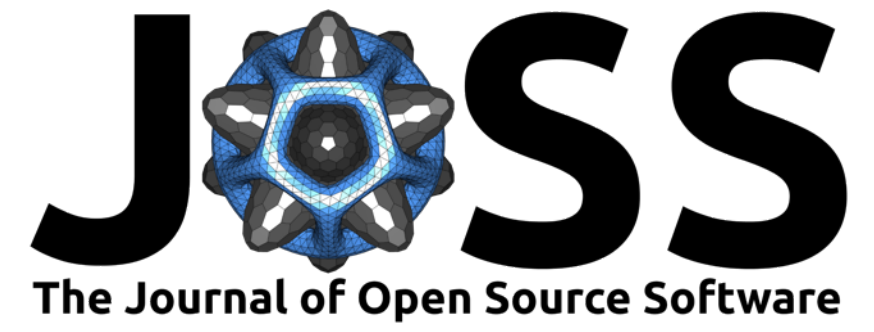

# HiGP: A high-performance Python package for Gaussian Processes

**Hua Huang** [1*], **Tianshi Xu** [2*], **Yuanzhe Xi** [2], and **Edmond Chow** [1¶]

**1** School of Computational Science and Engineering, Georgia Institute of Technology, USA **2** Department of Mathematics, Emory University, USA ¶ Corresponding author * These authors contributed equally.

## Summary

Gaussian Processes (GPs) (Rasmussen & Williams, 2005) are flexible, nonparametric Bayesian models widely used for regression and classification because of their ability to capture complex data patterns and quantify predictive uncertainty. However, the $\mathcal{O}(n^3)$ computational cost of kernel matrix operations poses a major obstacle to applying GPs at scale. HiGP is a high-performance Python package designed to overcome these scalability limitations through advanced numerical linear algebra and hierarchical kernel representations. It integrates $\mathcal{H}^2$ matrices to achieve near-linear complexity in both storage and computation for spatial datasets, supports on-the-fly kernel evaluation to avoid explicit storage in large-scale problems, and incorporates a robust Adaptive Factorized Nyström (AFN) preconditioner (Zhao et al., 2024) that accelerates convergence of iterative solvers across a broad range of kernel spectra. These computational kernels are implemented in C++ for maximum performance and exposed through Python interfaces, enabling seamless integration with modern machine learning workflows. HiGP also includes analytically derived gradient computations for efficient hyperparameter optimization, avoiding the inefficiencies of automatic differentiation in iterative solvers. By serving as a reusable numerical engine, HiGP complements existing GP frameworks such as GPJax (Pinder & Dodd, 2022), KeOps (Charlier et al., 2021), and GaussianProcesses.jl (Fairbrother et al., 2022), providing a reliable and scalable computational backbone for large-scale Gaussian Process regression and classification.

## Gaussian Processes

For training points $\mathbf{X} \in \mathbb{R}^{n\times d}$, a noisy training observation set $\mathbf{y} \in \mathbb{R}^n$, and testing points $\mathbf{X}_* \in \mathbb{R}^{m\times d}$, a standard GP model assumes that the noise-free testing observations $\mathbf{y}_* \in \mathbb{R}^m$ follow a joint Gaussian distribution that depends on a set of parameters, including scale $f$, noise level $s$, and kernel parameters $l$. The GP model finds the optimal parameters $\Theta := (s, f, l)$ by minimizing the negative log marginal likelihood:

$$L(\Theta) = \frac{1}{2}\left(\mathbf{y}^\top \widehat{\mathbf{K}}^{-1}\mathbf{y} + \log|\widehat{\mathbf{K}}| + n\log 2\pi\right),$$

where $\widehat{\mathbf{K}}$ denotes the regularized kernel matrix. An optimization process usually requires the gradient of $L(\Theta)$:

$$\frac{\partial L}{\partial \theta} = \frac{1}{2}\left(-\mathbf{y}^\top \widehat{\mathbf{K}}^{-1}\frac{\partial \widehat{\mathbf{K}}}{\partial \theta}\widehat{\mathbf{K}}^{-1}\mathbf{y} + \mathrm{tr}\left(\widehat{\mathbf{K}}^{-1}\frac{\partial \widehat{\mathbf{K}}}{\partial \theta}\right)\right), \quad \theta \in \Theta.$$

Using preconditioned iterative methods with preconditioner $\mathbf{M} \approx \widehat{\mathbf{K}}$ is a common option (Aune et al., 2014; Chen et al., 2023; Hensman et al., 2013; Pleiss et al., 2018; Wenger et al., 2022;

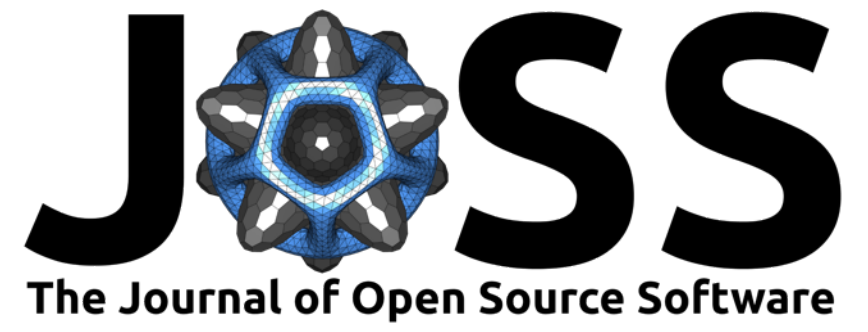

Wilson et al., 2015; Zhang et al., 2024). In this approach, $\widehat{\mathbf{K}}^{-1}\mathbf{y}$ is approximated via the preconditioned conjugate gradient (PCG) method (Saad, 2003). To handle the logarithmic determinant and trace terms, they are first rewritten as

$$\log|\widehat{\mathbf{K}}| = \log|\mathbf{M}| + \log|\mathbf{M}^{-1/2}\widehat{\mathbf{K}}\mathbf{M}^{-1/2}|, \tag{1}$$

$$\text{tr}(\widehat{\mathbf{K}}^{-1}\frac{\partial\widehat{\mathbf{K}}}{\partial\theta}) = \text{tr}\left(\mathbf{M}^{-1}\frac{\partial\mathbf{M}}{\partial\theta}\right) + \text{tr}\left(\widehat{\mathbf{K}}^{-1}\frac{\partial\widehat{\mathbf{K}}}{\partial\theta} - \mathbf{M}^{-1}\frac{\partial\mathbf{M}}{\partial\theta}\right). \tag{2}$$

The second component of each new expression is then estimated using the stochastic Lanczos quadrature (Ubaru et al., 2017) and the Hutchinson estimator (Hutchinson, 1989; Meyer et al., 2021), respectively.

# Statement of Need

The Gaussian Process (GP) community has advanced rapidly in recent years, developing scalable inference frameworks and more efficient kernel representations. Modern libraries such as GPyTorch (Gardner et al., 2018), GPflow (Matthews et al., 2017; van der Wilk et al., 2020), GPJax (Pinder & Dodd, 2022), KeOps (Charlier et al., 2021), and GaussianProcesses.jl (Fairbrother et al., 2022) leverage GPUs and automatic differentiation to perform GP inference efficiently on moderately large datasets. Concurrently, new algorithms, including preconditioned optimization methods (Wenger et al., 2022), alternating-projection solvers (Wu et al., 2024), GPU-accelerated Vecchia approximations for spatial data (Pan et al., 2024), robust relevance-pursuit inference (Ament et al., 2024), and latent Kronecker formulations for structured covariance matrices (Lin et al., 2025), have further improved the scalability and robustness of GP models. Yet, most existing frameworks emphasize modeling flexibility and seamless integration with autodiff ecosystems, rather than optimizing the low-level numerical routines that dominate runtime for very large or ill-conditioned kernel systems. HiGP is designed to address this computational gap by focusing on the numerical core of GP inference. It provides robust, scalable, and hardware-efficient implementations of kernel algebra, preconditioned iterative solvers, and gradient computations, offering three primary contributions.

Firstly, HiGP addresses the efficiency of MatVec, the most performance-critical operation in iterative methods. For large 2D or 3D datasets, the dense kernel matrix is compressed into a $\mathcal{H}^2$ matrix (Hackbusch et al., 2000; Hackbusch & Börm, 2002) in HiGP, resulting in $\mathcal{O}(n)$ storage and computation costs. For large high-dimensional datasets, HiGP computes small kernel matrix blocks on-the-fly and immediately uses them in MatVec and discards them, which allows HiGP to handle extremely large datasets with a moderate memory size.

Secondly, HiGP uses iterative solvers with the newly proposed AFN preconditioner (Zhao et al., 2024), which is designed for robust preconditioning of kernel matrices. Experiments demonstrate that AFN can significantly improve the accuracy and robustness of iterative solvers for kernel matrix systems. Furthermore, AFN and $\mathcal{H}^2$ matrix computation rely on evaluating many small kernel matrices in parallel, which is easily handled in C++ but would incur large overhead in Python, making implementation in other libraries such as GPyTorch or GPFlow more challenging.

Lastly, HiGP uses accurate and efficient hand-coded gradient calculations. GPyTorch relies on the automatic differentiation (autodiff) provided in PyTorch to calculate gradients (Equation 2). However, autodiff can be inefficient and inaccurate for computing the gradient of the preconditioner, so we use hand-coded gradient calculations for better performance and accuracy.

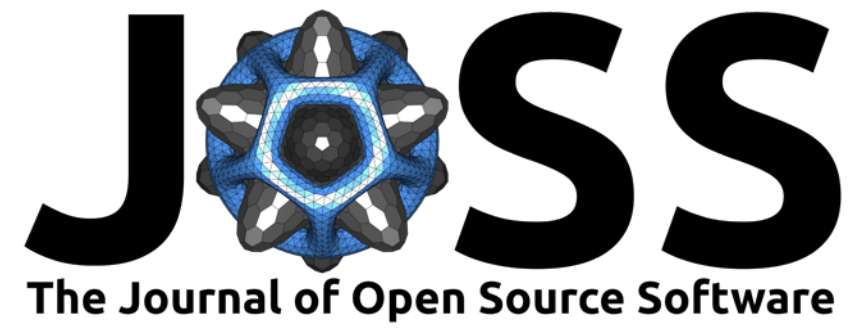

## Design and Implementation

We implemented HiGP in Python 3 and C++ with the goal of providing both a set of ready-to-use out-of-the-box Python interfaces for regular users and a set of reusable high-performance shared-memory multithreading computational primitives for advanced users. The HiGP C++ code implements all performance-critical operations. The HiGP Python code wraps the C++ units into four basic Python modules: `krnlmatmodule` for computing kernel matrices and its derivatives, `precondmodule` for PCG solver with AFN preconditioner, `gprproblemmodule` and `gpcproblemmodule` for computing the the loss and gradient for GP regression and classification. The two modules `gprproblemmodule` and `gpcproblemmodule` allow a user to train a GP model with any gradient-based optimizer.

We further implemented two high-level modules `GPRModel` and `GPCModel` using PyTorch parameter registration and optimizer to simplify the training and use of GP models. Listing 1 shows an example of defining and training a GP regression and using the trained model for prediction.

```
# Listing 1: HiGP example code of training and using a GPR model
gprproblem = higp.gprproblem.setup(data=train_x, label=train_y,
                                   kernel_type=higp.GaussianKernel)
model = higp.GPRModel(gprproblem)
optimizer = torch.optim.Adam(model.parameters(), lr=0.1)
for i in ranges(max_steps):
    loss = model.calc_loss_grad()
    optimizer.step()
params = model.get_params()
pred = higp.gpr_prediction(train_x, train_y, test_x,
                           higp.GaussianKernel, params)
```

We note that the HiGP Python interfaces (except for `GPRModel` and `GPCModel` models) are *stateless*. This design aims to simplify the interface and decouple different operations. A user can train and use different GP models with the same or different data and configurations in the same file.

## Numerical Experiments

We conducted numerical experiments on an Ubuntu 20.04 LTS machine with dual Intel Xeon Gold 6248R CPU (2x12 cores in total). We used PyTorch 2.8.0, GPyTorch 1.14, and HiGP version 2025.11.3 for the tests.

We tested two data sets from the UCI Machine Learning Datasets: the "Bike Sharing" and the "3D Road Network" data sets. We also tested three synthetic target functions from the Virtual Library of Simulation Experiments with randomly sampled data points: Rosenbrock, Rastrigin, and Branin. All datasets were normalized with Z-score normalization ($\mu = 0$, $\sigma = 1$) applied to both features and targets using statistics from the training set. The results represent averages over three independent runs for statistical reliability. Both HiGP and GPyTorch were configured with identical computational budgets to ensure a fair comparison. E2E tests use the following settings:

- Optimizer steps: 50
- CG iterations: 20 (training), 50 (prediction)
- Preconditioner/AFN rank: 10 (training), 100 (prediction)
- Optimizer: Adam with a learning rate of 0.01
- Precision: 32-bit floating point (FP32)

We first compared the end-to-end (E2E) accuracy and performance between HiGP and GPyTorch. Table 1 shows HiGP achieves equivalent GP accuracy compared to GPyTorch, and

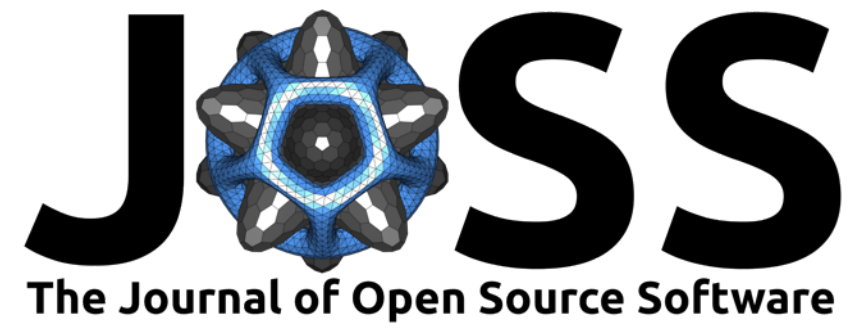

Table 2 shows HiGP has better performance than GPyTorch under fixed computational budget constraints and can handle some large datasets that GPyTorch cannot handle.

**Table 1:** HiGP accuracy tests on small data sets

| Dataset | n_train | Kernel | HiGP Mode | HiGP Final RMSE | GPyTorch Final RMSE |
|---|---|---|---|---|---|
| Bike | 3,000 | RBF | dense | 0.0285 | 0.0284 |
| Rosenbrock (5D) | 3,000 | Matern32 | dense | 0.0603 | 0.0658 |

**Table 2:** Performance comparison between HiGP and GPyTorch on large datasets

| Dataset | n_train | Kernel | HiGP Mode | HiGP Time (s) | GPyTorch Time (s) |
|---|---|---|---|---|---|
| Road3D | 50,000 | RBF | H2 | 191.2 | 14,739.3 |
| Road3D | 150,000 | RBF | H2 | 383.5 | — |
| Branin | 50,000 | RBF | H2 | 132.2 | — |
| Branin | 150,000 | RBF | H2 | 262.9 | — |
| Rastrigin (2D) | 30,000 | Matern32 | H2 | 100.6 | 278.2 |
| Rastrigin (20D) | 30,000 | Matern32 | on-the-fly | 198.5 | 275.6 |
| Rosenbrock (2D) | 30,000 | RBF | H2 | 82.2 | 231.3 |
| Rosenbrock (20D) | 30,000 | RBF | on-the-fly | 190.8 | 230.6 |

We also tested the parallel strong scaling ability of HiGP. Table 3 and Table 4 show HiGP has a good parallel scalability, and $\mathcal{H}^2$ matrix results match with the strong scaling results in H2Pack (Huang et al., 2020) since HiGP uses the same $\mathcal{H}^2$ matrix parallel algorithms as H2Pack.

**Table 3:** HiGP strong scaling performance test results with the 2D Rastrigin data set, Matern32 kernel, and using $\mathcal{H}^2$ matrix method

| Cores | Training | | | Inference | | |
|---|---|---|---|---|---|---|
| | Time (s) | Speedup | Efficiency | Time (s) | Speedup | Efficiency |
| 1 | 1480.51 | 1.00x | 100% | 30.62 | 1.00x | 100% |
| 2 | 759.45 | 1.95x | 97% | 15.73 | 1.95x | 97% |
| 4 | 399.82 | 3.70x | 93% | 8.33 | 3.67x | 92% |
| 8 | 215.49 | 6.87x | 86% | 4.47 | 6.85x | 86% |
| 16 | 133.78 | 11.07x | 69% | 2.86 | 10.71x | 67% |

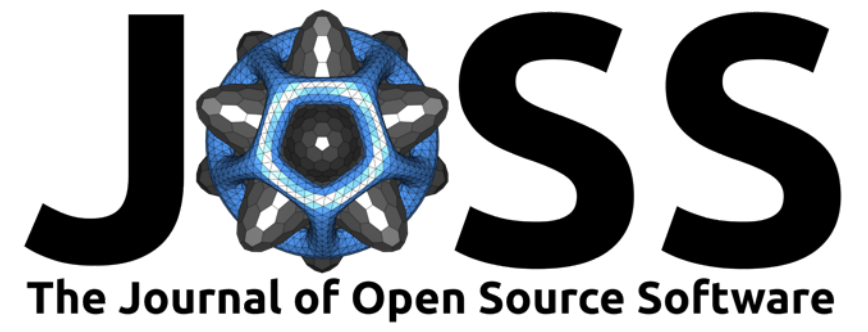

Table 4: HiGP strong scaling performance test results with the 20-D Rastrigin data set, Matern32 kernel, and using dense/on-the-fly method

| Cores | Training | | | Inference | | |
|---|---|---|---|---|---|---|
| | Time (s) | Speedup | Efficiency | Time (s) | Speedup | Efficiency |
| 1 | 2599.17 | 1.00x | 100% | 16.57 | 1.00x | 100% |
| 2 | 1411.49 | 1.84x | 92% | 9.02 | 1.84x | 92% |
| 4 | 744.72 | 3.49x | 87% | 5.03 | 3.29x | 82% |
| 8 | 398.91 | 6.52x | 81% | 2.80 | 5.92x | 74% |
| 16 | 229.92 | 11.30x | 71% | 1.76 | 9.41x | 59% |

## Acknowledgements

The research of Y. Xi is supported by NSF award DMS-2338904. The research of H. Huang and E. Chow is supported by NSF award OAC-2003683.

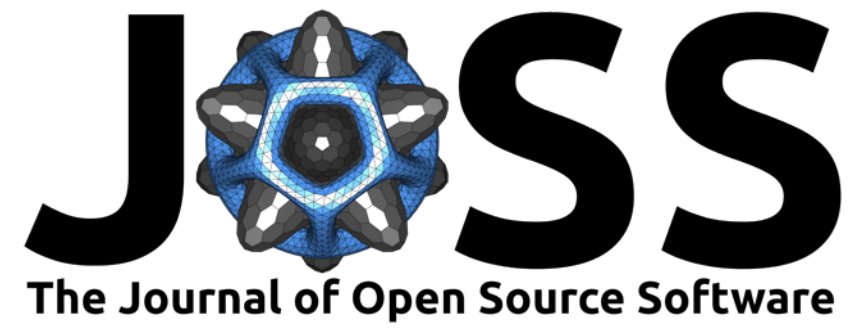